\documentclass[conference]{IEEEtran}
\IEEEoverridecommandlockouts
\usepackage{hyperref}
\usepackage{color}
\usepackage{xcolor}
\usepackage{tikz,tikz-qtree}
\usepackage{bbding} 
\usepackage{multirow}
\usepackage{amsmath}
\usepackage{amssymb}
\usepackage{mathtools}
\usepackage{amsthm}
\usepackage{amsfonts}
\usepackage{booktabs}
\usepackage{bbm}
\usepackage[numbers,sort&compress]{natbib}
\usepackage{makecell}
\usepackage{array}
\newcolumntype{F}[1]{%
    >{\raggedright\arraybackslash\hspace{0pt}}p{#1}}%
\newcolumntype{T}[1]{%
    >{\centering\arraybackslash\hspace{0pt}}p{#1}}%

\theoremstyle{plain}
\newtheorem{theorem}{Theorem}[]
\theoremstyle{definition}
\newtheorem{definition}[theorem]{Definition}
\AtBeginDocument{%
  \providecommand\BibTeX{{%
    \normalfont B\kern-0.5em{\scshape i\kern-0.25em b}\kern-0.8em\TeX}}}

\newcommand{\eg}[0]{{e.g.}}
\newcommand{\ie}[0]{{i.e.}}
\newcommand{\etal}[0]{{et al.}}

\begin{document}

\pagenumbering{arabic}  
\pagestyle{plain}

\title{Privacy of Autonomous Vehicles:
\\Risks, Protection Methods, and Future Directions}

\author{Chulin Xie$^1$, Zhong Cao$^2$, Yunhui Long$^1$, Diange Yang$^2$, Ding Zhao$^3$, Bo Li$^1$    \\
{\small$^1$University of Illinois Urbana-Champaign   $^2$Tsinghua University  $^3$Carnegie Mellon University} \\
{\footnotesize \texttt{\{chulinx2, ylong4,lbo\}@illinois.edu} \quad \texttt{\{caozhong, ydg\}@tsinghua.edu.cn}  \quad \texttt{dingzhao@cmu.edu}} \\
}

\maketitle

\begin{abstract}

Recent advances in machine learning have enabled its wide application in different domains, 
and one of the most exciting applications is autonomous vehicles (AVs), which have encouraged the development of several ML algorithms from perception to prediction to planning.
However, training AVs usually requires a large amount of training data collected from different driving environments (e.g., cities) as well as different types of personal information (e.g., working hours and routes). Such collected large data, treated as the ``new oil" for ML in the data-centric AI era, usually contains a large amount of privacy-sensitive information which is hard to remove or even audit.
Although existing privacy protection approaches have achieved certain theoretical and empirical success, there is still a gap when applying them to real-world applications such as autonomous vehicles.
For instance, when training AVs, not only can \textit{individually} identifiable information reveal privacy-sensitive information, but also \textit{population-level} information such as road construction within a city, and \textit{proprietary-level} commercial secrets of AVs. 
Thus, it is critical to revisit the frontier of privacy risks and corresponding protection approaches in AVs to bridge this gap.
Following this goal, in this work, we provide a new taxonomy for privacy risks and protection methods in AVs, and we categorize privacy in AVs into three levels: \textit{individual}, \textit{population}, and \textit{proprietary}.
We explicitly list out recent challenges to protect each of these levels of privacy, summarize existing solutions to these challenges, discuss the lessons and conclusions, and provide potential future directions and opportunities for both researchers and practitioners.
We believe this work will help to shape the privacy research in AV and guide the privacy protection technology design.
\end{abstract}

\section{Introduction}
In the past years, autonomous vehicles (AV) have encouraged a number of learning algorithms ranging from object detection~\cite{ess2010object, michaelis2019benchmarking} to 3D recognition~\cite{chen2017multi} and from prediction~\cite{mozaffari2020deep} to control and reinforcement learning (RL)~\cite{kabzan2019learning, aradi2020survey}, making it one of the most exciting machine learning real-world applications.   
Training AV models usually require a large amount of data collected from different driving environments such as cities and urban areas. In addition, different types of personal information, such as age, gender, and working time are also collected to help improve autonomous driving performance. It is obvious that such massive data collected contains sensitive private information that needs to be protected~\cite{fernandez2021trustworthy}. According to a recent user study~\cite{bloom2017self}, more than 50\% of participants believed that ``capturing images,'' ``aggregating and storing information'', and ``continuous analysis'' are very likely to happen in AVs, leading to great privacy risks. 
Moreover, more than 50\% of participants are uncomfortable with the secondary analysis of information such as recognition, identification, and tracking of their vehicles~\cite{bloom2017self}. 
As a result, given the wide data collection range and diverse trained model types 
(e.g., recognition and RL models)
, it is urgent to explore the detailed data types required for AV and thus build up a systematic and comprehensive understanding of privacy in AV, including the privacy risks,  potential protection approaches, and the connections with existing generic privacy research,  before the large deployment of AV. 

In the meantime,  different privacy risks (attacks) have been recognized in different standard machine learning (ML) models in the current data-driven era, 
such as the \textit{membership inference} attacks~\cite{shokri2017membership} that try to predict whether a data point belongs to the training set, and the \textit{model inversion} attacks~\cite{Fredrikson2015ccs} that aim to recover part of the training set (e.g., human faces).   
More recently,
advanced model inversion attack~\cite{hitaj2017deep} has been proposed against federated learning (FL)~\cite{mcmahan2017communication} systems, where users collaboratively train a model through a parameter server while keeping their raw data locally.
Given that companies such as Google have applied FL in real life, such as through mobile devices~\cite{mcmahan2017communication}, these privacy attacks~\cite{hitaj2017deep} can cause serious privacy concerns in practice.
In addition, large-scale language models are also found to leak private information, such as social security numbers and credit card numbers~\cite{carlini2019secret}, which again raises privacy concerns for different common services, such as  email auto-completion.
\begin{figure}[t]
	\centering
	\includegraphics[width=1\linewidth]{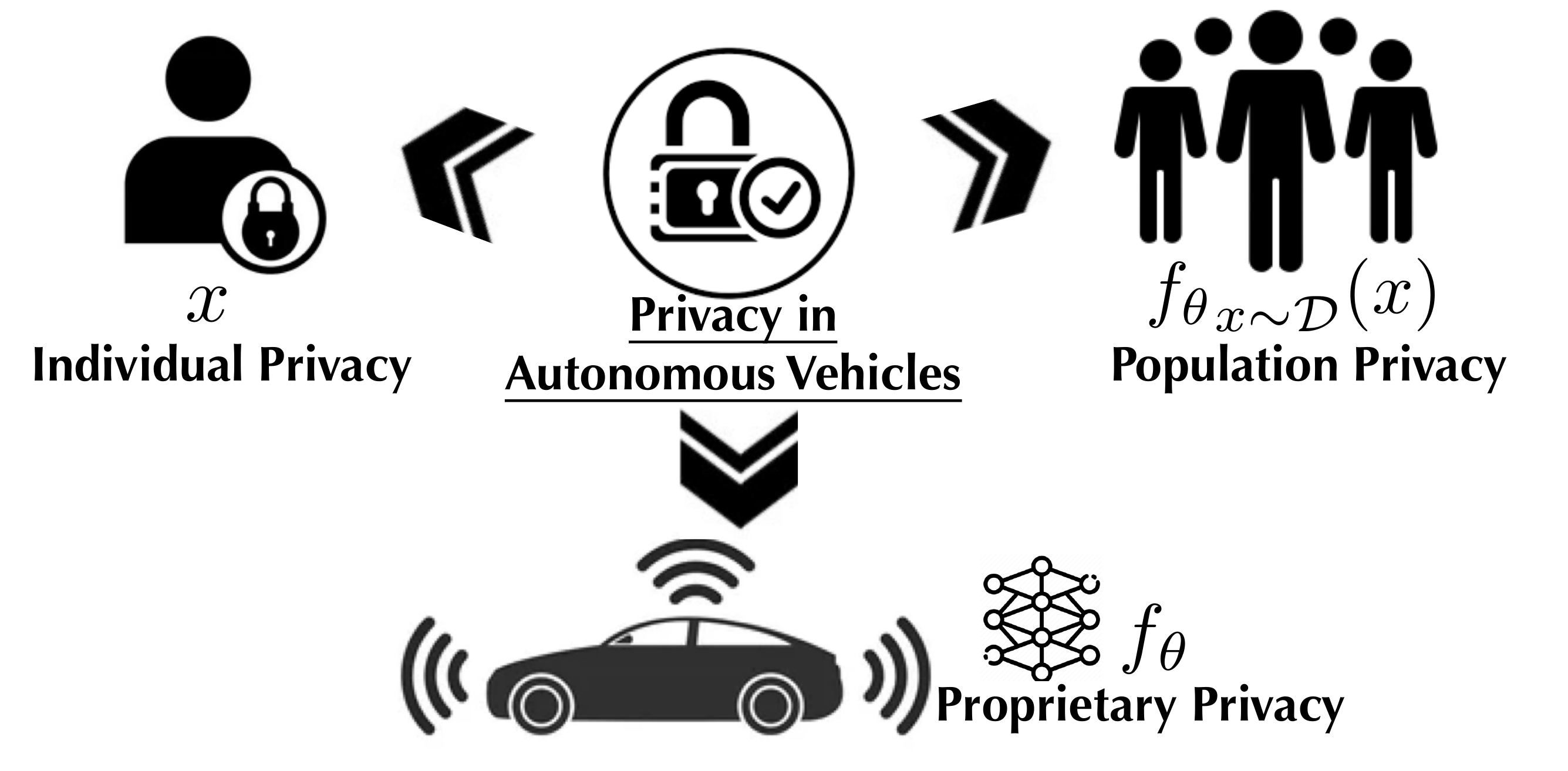}
	\caption{\textbf{Overview of privacy in autonomous vehicles}.}
	\label{fig-overview}
	\vspace{-6.4mm}
\end{figure}

However, although privacy risks universally exist in different domains,
the privacy-sensitive information and corresponding attack/protection strategies are different in each domain. For instance, population-level information such as the city views
is usually privacy sensitive~\cite{glancy2012privacy} in AVs, whereas population-level information is usually non-sensitive in other standard single ML model settings. 
Thus, it is of great importance to revisit the \textit{unique} privacy properties in AV~\cite{bloom2017self}, and thus design corresponding protection mechanisms in practice. 

Targeting the unique privacy properties and challenges in AV, in this work, we aim to summarize the privacy risks and protection strategies in the AV domain, so as to encourage the principled design of protection mechanisms.
In particular, we provide a novel taxonomy of privacy in AVs and categorize it into three categories: \textit{individual privacy}, \textit{population privacy}, and \textit{proprietary privacy} as shown in Figure~\ref{fig-overview}. 
For \textit{individual privacy}, we aim to discuss individual private information such as collected individual face images and working schedules, which follow similar privacy notions to existing privacy research. For \textit{population privacy}, the goal is to protect certain population-level statistics. 
Finally, since a range of ML models would be trained for AVs such as object recognition and decision-making models, the problem of protecting model parameters to avoid attacks such as model stealing leads to \textit{proprietary privacy}.


Regarding privacy protection, different
\textit{regulations} have been promulgated to protect private information, such as the Privacy Rule of the Health Insurance Portability and Accountability Act (HIPAA) of 1996 (when disclosing medical records)~\cite{nosowsky2006health}, the Federal Rules of Civil Procedure (when disclosing court records)~\cite{tobias1988public}, the European General Data Protection Regulation (GDPR)~\cite{regulation2018general}, and the California Consumer Privacy Act (CCPA)~\cite{ccpa2018}.
In the meantime, a line of \textit{algorithmic approaches} have also been explored to protect data privacy in ML during the last decades, including privacy-preserving \textit{learning} algorithms and 
privacy-preserving data \textit{generative} models. 
There is a long history of research to design specific \textit{privacy-preserving learning algorithms} 
based on different privacy notions, such as $k$-anonymity~\cite{sweeney2002k}, $l$-diversity~\cite{machanavajjhala2007diversity}, $t$-closeness~\cite{li2007t} and the gold standard $(\epsilon,\delta)$-differential private (DP)~\cite{dwork2006calibrating, dwork2008differential,dwork2014algorithmic}.
Differentially private ML models for image data~\cite{abadi2016deep,papernot2018scalable}, large language models~\cite{mcmahan2017learning, li2021large,yu2021differentially}, and reinforcement learning~\cite{wang2019privacy,ma2019differentially} have all been developed respectively. 
To {generate} large-scale data with  privacy guarantees for \textit{any} downstream tasks, several \textit{privacy-preserving data generative models} have been proposed, such as the differential private generative adversarial networks (GAN)~\cite{xie2018differentially,chen2020gs,takagi2021p3gm}.
Some of these protection approaches can be adapted to the AV domain, while new protection mechanisms also need to be designed according to the additional privacy challenges in AV. 

In this work, we will introduce existing privacy risks and attacks, together with the corresponding protection algorithms, regarding individual privacy (Section~\ref{section:individual_privacy}), population privacy (Section~\ref{section:population_privacy}), and proprietary privacy (Section~\ref{section:proprietary_privacy}). We will also discuss challenges and future research directions of privacy in AV in Section~\ref{section:discussion_future_directions}.

\begin{figure}[t]
\centering
\includegraphics[width=1\linewidth]{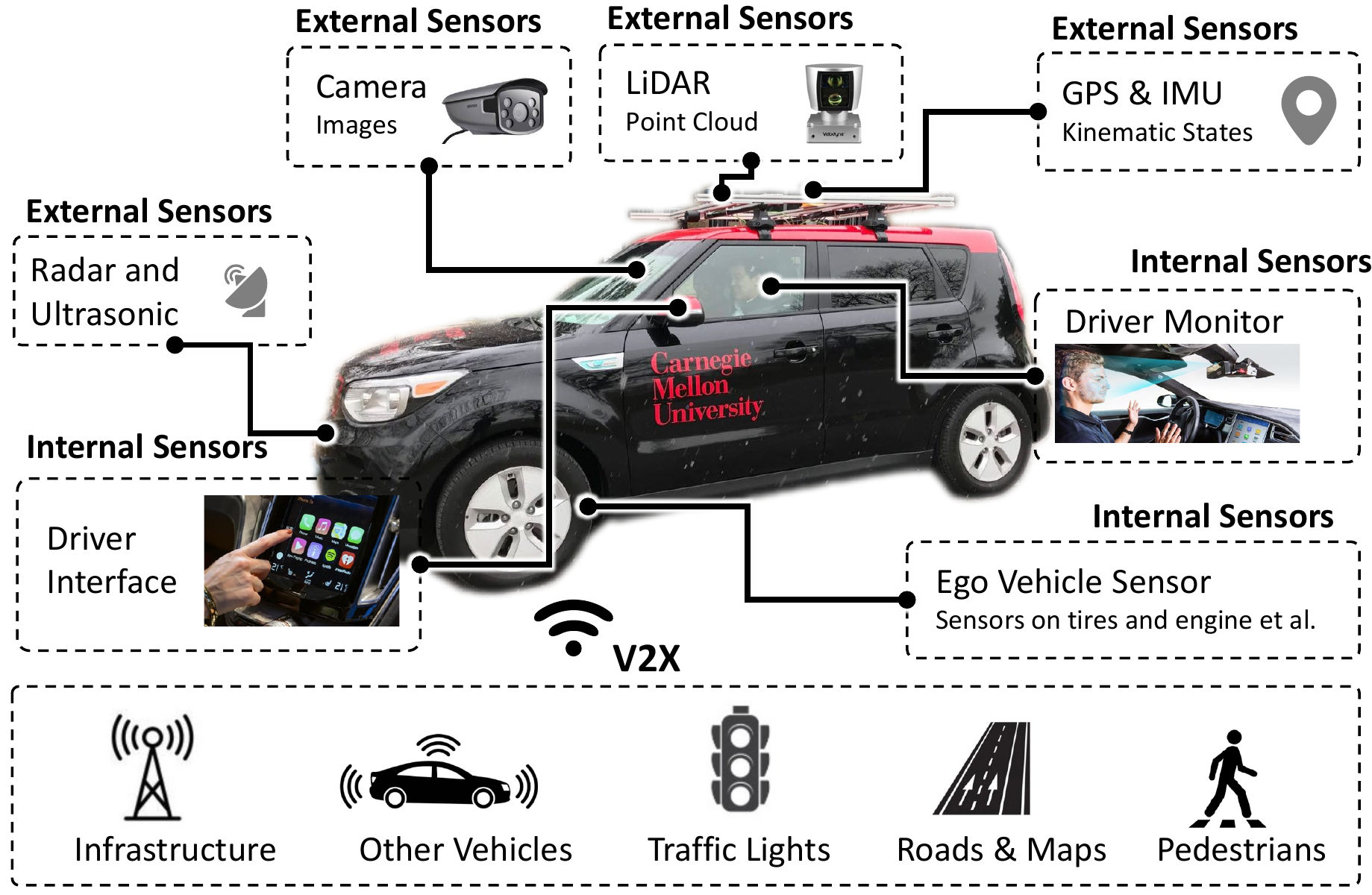}
\caption{\small \textbf{Data Types in Autonomous Vehicles.}}
\label{fig:Section2/DataTypes}
\vspace{-2mm}
\end{figure}

\section{Data-Centric AV and Privacy Risks}
\label{sec:data-centric-av}

A typical autonomous vehicle is expected to drive to the destination efficiently and safely.
It has to continuously collect the necessary information from the surrounding environment and the passengers,
which may cause privacy leakage.

\subsection{Data Types in Autonomous Vehicles}

AVs collect the data on the surrounding environment, passengers, and roadside facilities \cite{yang2018intelligent}, and we categorize AV data into three categories: external sensing data, internal sensing data, and V2X data according to their usage environments (Fig. \ref{fig:Section2/DataTypes}).

\underline{(1) External sensing data}
is collected in external sensors, including LiDAR, Radars, Cameras, and Ultrasonic, to monitor the external environment.
We focus on the aspects of the sensors most likely related to privacy, and a more comprehensive discussion of these sensors can be found in \cite{campbell2018sensor}.
Specifically, \textbf{Lidar} detects the range information of the surroundings and generates point clouds \cite{zermas2017fast}, 
which contain environmental information (e.g., relative position and reflection intensity) in 360-degree or in specific directions~\cite{li2021towards}.
\textbf{Cameras \cite{sivaraman2013looking}} record images/video of the surroundings. 
An AV equips different cameras for a variety of functions \cite{sun2006road}, e.g., front view camera for car-following, side view camera for lane-change, and rear view camera for parking.
Cameras may access rich environmental information, often more than the requirements of the basic driving function. 
\textbf{Radar \cite{bilik2019rise} and Ultrasonic \cite{xu2018analyzing}} 
measure the range and bearing angle of surrounding objects, 
but their outputs usually have a much lower resolution than LiDAR.

\underline{(2) Internal sensing data} is collected to monitor the internal events inside the cabin.
The most critical sensors for the kinematic states are GPS and the Inertial Measurement Unit (IMU) \cite{sukkarieh1999high}, where GPS collects the position of the vehicle on the earth while IMU tracks the velocity and acceleration.
\textbf{Ego vehicle sensors} collect the kinematic states of the ego vehicle (e.g., location, speed, and acceleration) as well as the states of vehicle components (e.g., tires  \cite{velupillai2007tire} and engines \cite{song2002vehicle}).
The kinematic states can represent the driving trajectory with time accumulation.
\textbf{Passenger Interface \cite{faas2020external}} interacts with passengers via onboard Human-Machine Interaction system~\cite{gorecky2014human} or cellphone connections.
Passengers can input their requirements and preferences, such as the destinations, route preferences, or temperatures.
Some sensors also collect the physiological states to provide more specific services, such as fatigue testing \cite{gao2015evaluating} and cell phone usage \cite{seshadri2015driver}.
Such systems can continuously monitor passengers.
Companies or automakers also launch operating systems that allow third-party apps such as Blackberry QNX OS~\cite{chemerkin2012vulnerability}, 
which opens up more channels for collecting private user information.

\underline{(3) V2X communication data} is received via vehicle to everything (V2X) communication.
V2X can be roughly divided into roadside infrastructure, road users, and AV service providers.
The data from \textbf{roadside infrastructure (V2I)} includes the traffic light states, traffic rules, etc. 
Cameras also detect these data, but it is easier and more accurate to receive the information from the V2I system.
Some V2I systems may also broadcast the suggested driving speed and reminders of the accident ahead \cite{chu2021cloud}.
The data from \textbf{road users (V2V)} includes the surrounding road users' information (e.g., kinematic states).
Some research areas including vehicle platoon cooperative control and cooperative perception strongly rely on the V2V system \cite{zhu2020v2v}. 
Vehicle drivers may also broadcast their intention to enhance driving safety \cite{li2021planning}.
The data from \textbf{AV service provider} includes traffic density \cite{9357917} and HD map, where HD map consists of the lanes, traffic rules, and some static objects \cite{jiang2019flexible} in the surroundings. 
It provides more accurate information, thus significantly improving driving performance.
However, it also provides rich information about the city, which may bring privacy risks.
\subsection{Data Usages of Autonomous Vehicles}
We will briefly describe the usages of data in an AV system based on ``perception - planning - control'', which is the default setting for popular open-source AV systems such as Baidu Apollo \cite{fan2018baidu}, Autoware \cite{kato2018autoware}, and CMU's Tartan \cite{urmson2007tartan}.

\underline{(1) Data usages for perception} refers to extracting the key surrounding information by the external and internal sensors or vehicle communication systems. 
\textit{First}, from sensor data, AVs perceive objects (e.g., pedestrians, vehicles, cyclists, and other uncommon objects), roads (e.g., lanes, road edges, intersections) 
and traffic rules (e.g., traffic lights, speed, limits and stop signs), where deep neural networks (DNNs), e.g., YOLO \cite{redmon2018yolov3} and PSPNet \cite{zhao2017pyramid}, are widely adopted for object detection.
\textit{Second}, the received data from V2X communication enables blind spot observation~\cite{thakurdesai2021autonomous}, redundant perception~\cite{kim2014multivehicle},  cooperative perception~\cite{kim2014multivehicle}, and surrounding views perception~\cite{9357917}.

\underline{(2) Data usages for planning and control} means
planning
the driving trajectory using environment information. 
\textit{First}, with sensor data, the ego vehicle predicts the future motions of surrounding objects with DNNs~\cite{gao2020vectornet,liang2020learning} to avoid collisions, and  plans the trajectories considering the driver preference (e.g., conservative or aggressive driving style). 
\textit{Second}, with V2X communication, kinematic states of other vehicles and other drivers' intentions empower platoon or multi-agent planning \cite{petrillo2018adaptive} for higher efficiency, while received HD map enables routing and shared AV services such as Uber and DIDI.

In general, all the current data usages are necessary for an AV system.
However, they may have significant privacy risks.

\subsection{Data Usage on Different Privacy Levels}
Previous sections introduce ``\textit{how to collect and use the data}'' in the AV system, 
whereas privacy concerns focus on ``\textit{what information is contained in these data}''.
Thus, to better introduce privacy risks and defending methods, we classify AV privacy into three levels: individual privacy, population privacy, and proprietary privacy.

\underline{(1) Individual privacy} is about the information leakage of people using the AV service or involved in an AV system, such as AV owners, passengers, and pedestrians.
AV poses a threat to individual-level privacy due to the rich amount of information collected by the sensors, the data being accessible to a broader audience through AV-related services, and the leakage of individual data through ML models. 
This is the most common privacy concern around the use of AVs.
Most information related to this kind of privacy is collected by single-vehicle sensors. 

\underline{(2) Population privacy} involves sensitive information from a group of data, causing population-level privacy risks. 
The data collected from the same city or area may build city views and leak information about infrastructures and sensitive locations.
These data can also be used to form a ubiquitous sensor network, which serves as a widespread surveillance system.
Since population-level information requires data on a large scale, it may be contained in a connected vehicle network or vehicle cloud.

\underline{(3) Proprietary privacy} indicates the information leakage of companies' proprietary around AVs,
which usually contains company secrets to developing an AV, e.g., the AV systems' algorithms, parameters, models, and frameworks. 
Among them, some ML models may be confidential for AV companies and should be carefully protected.
Some apps in AVs or connected vehicle networks may easily assess this information, causing privacy concerns for some big companies.
This privacy mainly uses connected vehicle networks and vehicle cloud information. 
In the following sections, we will introduce key concerns and protections for the three levels of privacy.

\begin{table}[t]
\centering
\scriptsize
\caption{\textbf{Different Levels of Privacy  in AVs.}}
 \vspace{-2mm}
\resizebox{\columnwidth}{!}{%
\begin{tabular}{lll}
\toprule
\textbf{Privacy Level}  & \textbf{Examples of Related AVs Data and Privacy Risks}  \\ 
\midrule
\multirow{6}{*}{\makecell[l]{Individual Privacy\\ (Individual user $x$)}} 
& Information about AV owners, passengers, and pedestrians: \\ 
& - single-vehicle data accessible by a broader audience~\cite{kopanaki2012framework, gao2017identifying,de2021using, shang2012user, taeihagh2019governing, privacy_invading_eye_2021} \\ 
& - linkage with external data sources~\cite{lim2018autonomous,punhani2020privacyberkeley}\\ 
& - membership inference in ML training datasets~\cite{shokri2017membership,long2020pragmatic,salem2019ml,truex2019demystifying,yeom2018privacy,choquette2021label,hayes2019logan,liu2019performing,pyrgelisknock} \\ 
& - model inversion in ML training datasets~\cite{Fredrikson2015ccs, yang2019neural, zhang2020secretrevealer} \\
& - model memorization in ML training datasets~\cite{song2017machine,carlini2019secret} \\
\hline
\multirow{3}{*}{\makecell[l]{Population Privacy \\ (Population  $f_\theta(x)$)}}
& Population-level data aggregated across different vehicles: \\
& - road and city views conditions~\cite{garg2018ubiquitous} \\
& - widespread surveillance~\cite{glancy2012privacy,calavia2012semantic, thomas2015combining} \\ 
\hline 
\multirow{4}{*}{\makecell[l]{Proprietary Privacy \\ (Model information  $f_\theta$)}}
& Proprietary information owned by the AV manufacturer: \\
& - hardware information~\cite{reiss2012obfuscatory} \\
& - model architecture~\cite{yan2020cache, hong2019how, liu2020ganred, hu2020deepsniffer, batina2019csi, duddu2018stealing, yu2020deepem, hua2018reverse} \\ 
& - model parameters and hyper-parameters~\cite{milli2019model, tramer2016stealing, chandrasekaran2020exploring, krishna2019thieves, carlini2020cryptanalytic, rolnick2020reverse, shen2022model, chen2021stealing, wang2018stealing} \\
\bottomrule
\end{tabular}
}
\vspace{-6mm}
\label{tab:cloud_privacy_concept}
\end{table}

\section{Individual Privacy in Autonomous Vehicles}
\label{section:individual_privacy}
\textbf{Overview.} 
\emph{Individual privacy} mainly focuses on protecting the privacy of individual data, such as the recorded human face or voice within the car and the individual driving trajectory.
In general,  differential privacy (DP)~\cite{dwork2014algorithmic} is widely used to protect individual privacy by ensuring that the information about \emph{individuals} can be hidden in the \textit{population}. 
The following definition formally describes this privacy guarantee. 
\begin{definition}[$(\epsilon, \delta)$-DP]
A randomized algorithm $\mathcal{M}$ with domain $\mathbb{N}^{|\mathcal{X}|}$ is $(\epsilon,\delta)$-\textit{individual} differentially private if for all
$\mathcal{S} \subseteq {\text{Range}(\mathcal{M}})$ 
and for any neighboring datasets $D$ and $D'$, we have
$\Pr[\mathcal{M}(D) \in \mathcal{S}] \leq \exp(\epsilon)\Pr[\mathcal{M}(D')\in \mathcal{S}] + \delta.$
\end{definition}
By defining neighboring datasets $D$ and $D'$ as two datasets differing only by an individual data sample $x$, DP aims to protect individual privacy by ensuring that with or without each data sample, the model prediction would be similar, so as to protect the membership information of instances.

In AV, individual privacy often refers to the privacy of individuals that interact with an AV, such as owners, passengers, and pedestrians.   
In particular, the privacy concerns are raised from two aspects: (1) direct sensitive data leakage (Section~\ref{sec:individual-data}), and (2) different privacy attacks against trained ML models in AV to infer individual sensitive information (Section~\ref{sec:individual-model}).
For instance, as discussed in section~\ref{sec:data-centric-av}, each autonomous vehicle collects extensive data from individual users, such as location traces, facial identity, user behavior data, and voice data. These collected data would lead to serious privacy concerns, and we will illustrate them and corresponding protection strategies (Section~\ref{sec:individual-protection}).

\subsection{Privacy Risks on Individual-Level}
Here we will introduce the potential privacy leakage on the individual level, as well as potential privacy attacks, given a trained machine learning model or database.
According to a user survey~\cite{bloom2017self}, the top privacy concerns on AVs include location tracking, individual tracking, and individual identification. 
In addition, the collected AV data in the database or cloud further increases these privacy risks due to the potential risks of internal and external data misuse. For example, Uber employees have been reported to stalk celebrities using the location data collected by the company~\cite{uber2021}, and an external data breach has caused the loss of location histories for 57 million riders and drivers~\cite{uberdatabreach2021}. 
Besides the leakage of user data, tracking pedestrians through external cameras also brings significant privacy risks for non-user individuals.

\subsubsection{Privacy Sensitive Data on Individual-Level}
\label{sec:individual-data}
We categorize the individual level data leakage in AV based on the \textit{data usage} in the machine learning life-cycle of AV, including 1) \textit{perception} and 2) \textit{planning and control}, which are related to different autonomous driving algorithms training, and 3) \textit{user experiences}, which are for improving personal driving and riding experience. 

\underline{Perception:}
 The data collected by different sensors in AV, such as cameras, radar, thermal imaging devices,  and light detection and ranging (LiDAR) devices, include 2D images, videos, audio, and 3D data of the surroundings for environment perception. 
Such perception data raises two privacy concerns, namely, \textit{identity disclosure} (\ie, identifying individuals in the dataset) and \textit{location privacy} (\ie, obtaining individuals’ spatial and temporal information).
    \textit{First}, leveraging   detection ML models, such as license plate detection~\cite{xie2018robustlicenseplate,yuan2016robustlicenseplate}, pedestrian face detection~\cite{Brkic2017,yang2016wider, hjelmaas2001face}, text detection~\cite{li2000automatic,ye2014text},
    one can detect the external entities from the perception data, which reveals the personal identities of external pedestrians. 
    \textit{Second}, the location of vehicles or pedestrians can be easily inferred from the surrounding street environment in perception data. Therefore, the collected perception data can be used to track any targeted external entities.

\underline{Planning and control:}
The data collected by AVs for planning and control purposes includes detailed GPS information, route, speed, and travel time, etc. 
    \textit{First}, by observing the collected data from an individual user over time, one can directly infer his or her daily activities. For example, an individual’s familial, religious, medical, or sexual details could be revealed by knowing his or her historical trips to psychiatrists, places of worship, hospitals, etc. It may increase the risk of being stalked, robbed, or attacked if the data is not secured. 
    \textit{Second}, by querying and mining the trajectory databases scored in the cloud, the attacker can perform privacy attacks such as  user identification, sensitive location, and sequential tracking privacy attacks~\cite{kopanaki2012framework}. 
    \textit{Third}, the trajectory models for different  applications (\eg, trajectory recommendation and trajectory classification) in AV systems are trained by individuals' mobility data, and by analyzing the model outputs, one may infer the sensitive properties of the training data. 
    For instance, 
    for trip recommendations, Shang~\etal~\cite{shang2012user} present a user-oriented trajectory search; such a model may disclose the frequently visited locations of the user by predicting the user's most likely next stopping point.
    Moreover, in trajectory classification, i.e., identifying who is the user of a trajectory, 
    recent works~\cite{gao2017identifying,de2021using}  present the RNN-based approach to identify and link trajectories to users via location-based trajectory embeddings.
    Such trajectory-user linkage (attack) raises great privacy concerns since the person's identity could be revealed by model inference with personal trajectory information.

\underline{User experience:}
Usually, AVs would collect data on drivers to improve user experiences.
For example, it may collect health data including eye movements, heart rates, and car accident-related data, are collected to monitor driver behaviors, which could contribute to safe driving recommendations and reminder systems. 
For instance, eye movements are used to warn the driver when they are dozing off ~\cite{MULDER2021105530}, drive gaze/attention maps obtained from driver eye movements can guide AV models and improve their performance and safety~\cite{xia2019drivereye}, and when there is a deviation in the user’s heart rhythm, the AV can warn the emergency services ~\cite{MULDER2021105530}. However, studies have shown that eye tracking technologies could be used to predict sensitive personal information, such as gender, age, cognitive disorders, mental and psychological illness, HIV/AIDS, etc.~\cite{privacy_invading_eye_2021}, and the collected heart rates can be used to infer the private health condition of the drivers which would affect their medical insurance.
Moreover, the collected user data may harass the drivers through tailored advertising and marketing strategies. For example, the data of EDRs, which ascertains the exact causes of accidents may be sold to third parties such as insurance companies and used against drivers~\cite{taeihagh2019governing}.

In addition, with \textit{external knowledge} and information sources, the collected individual AV data can further leak private information. Such external information sources include social networks, news articles, and chat logs. 
For instance, Lim~\etal~\cite{lim2018autonomous} show that pedestrians' geographical locations can be used for re-identification with the aid of a small amount of side information. 
Punhani~\etal~\cite{punhani2020privacyberkeley} extract license plates, pedestrian faces, and vehicle location information and then create a pipeline that performs a joint privacy attack by identifying quasi-identifiers. 
With such external knowledge, privacy-sensitive information is more likely to be leaked.

\subsubsection{Individual Privacy Attacks against ML Models in AV}
\label{sec:individual-model}
Given the collected individual privacy sensitive data mentioned above, we will next introduce how the trained ML models in AV would leak such information. In particular, given the trained ML models, various types of privacy attacks can be conducted. Such privacy attacks include \textit{membership inference attacks}, \textit{model inversion attacks}, and \textit{model memorization attacks}, which are summarized
in Table~\ref{tb:ind_attack_model}. 

\begin{table}[t]
    \centering
    \scriptsize
 \caption{\textbf{Individual Privacy: Risks and Privacy Attacks.}}
 \vspace{-2mm}
\resizebox{1\columnwidth}{!}{
\begin{tabular}{l l l l l l}
\toprule[1pt]
 \textbf{Attack Category}   &  \textbf{Attack target } & \textbf{Adversary’s Knowledge}   & \textbf{Attack Method} & \textbf{Data Type}  \\ 
\midrule[1pt]
\multirow{8}{*}{\makecell[l]{Membership \\ Inference}} & \multirow{5}{*}{\makecell[c]{Classification \\ models}}   &  \makecell[l]{Black-box model, \\input distribution,  model type}  &  shadow model   & \makecell[l]{location, image,\\ tabular data~\cite{shokri2017membership,long2020pragmatic}}    \\  \cmidrule{3-5}
& &  Black-box model  & confidence-thresholding & \makecell[l]{location, image,\\ text, tabular data~\cite{salem2019ml}}  \\  \cmidrule{3-5} 
& &  Black-box model& shadow dataset generation &  image,  tabular data~\cite{truex2019demystifying}  \\  \cmidrule{3-5} 
& &  \makecell[l]{Black-box model, \\average training loss}  &  threshold-based  & tabular data~\cite{yeom2018privacy}     \\  \cmidrule{3-5} 
& &   \makecell[l]{Black-box model, \\input distribution,  model type}   & \makecell[l]{shadow model, \\robustness evaluation}  & images, location, tabular~\cite{choquette2021label}   \\ \cmidrule{2-5} 
& \multirow{2}{*}{\makecell[c]{Generative \\ models}} &  White/Black-box model  &  \makecell[l]{confidence-thresholding, \\GAN} & images~\cite{hayes2019logan}  \\  \cmidrule{3-5} 
&  & Black-box model  & latent encoding & images~\cite{liu2019performing} \\ \cmidrule{2-5} 
&  \makecell[c]{Aggregated \\ statistics} & \makecell[l]{Black-box model, \\prior observation} & game-based procedure  & location~\cite{pyrgelisknock}  \\  \cmidrule{1-5}
\multirow{3}{*}{\makecell[l]{Model \\ Inversion}}  &  \multirow{2}{*}{\makecell[c]{Classification\\ models}} &  White/Black-box  model& optimization-based & image, tabular data~ \cite{Fredrikson2015ccs}    \\ \cmidrule{3-5}
 &   &   Black-box model, public data  & inversion model  &  image~\cite{yang2019neural} \\\cmidrule{3-5}
 &   &   \makecell[l]{White-box model, public data,\\ corrupted target input}  & GAN  &  image~\cite{zhang2020secretrevealer}  \\ \cmidrule{1-5}
 \multirow{2}{*}{\makecell[l]{Model \\ Memorization}}& \makecell[c]{Classification \\ models} &  White/Black-box model  & encoding  &  image, text~\cite{song2017machine}  \\  \cmidrule{2-5}
&  \makecell[c]{Generative \\ models} &   White-box model  & \makecell[l]{random sequences\\ insertion}  &  text~\cite{carlini2019secret}  \\ 
\bottomrule
\end{tabular}}
\vspace{-6mm}
  \label{tb:ind_attack_model}
\end{table}

\underline{Membership Inference Attacks:}
Membership inference (MI) attacks infer if a data record is part of the training set of an ML model. 
Inferring an individual’s membership in a dataset can have serious privacy implications. For instance, in  AV services, if an ML model is trained using trajectory records of a driver suffering from a sensitive medical condition (\eg, frequent cancer hospital visits),  a successful membership inference of the driver's trajectory record can reveal his/her sensitive medical conditions.

These types of attacks are often performed on machine learning as a service (MLaaS) platforms, where an attacker can query the prediction API of the target model and get the prediction results~\cite{shokri2017membership, hayes2019logan, salem2019ml}. Shokri~\etal ~\cite{shokri2017membership} create multiple shadow models
to imitate the target model's behavior, and then train an attack model based on the outputs of the shadow models to in turn differentiate the predictions of the target model trained with or without the sensitive record. 
Salem~\etal~\cite{salem2019ml} relax the assumptions of \cite{shokri2017membership} so that the attacker can still succeed when training a single shadow model with model architectures different from the target model and with datasets different from the original training dataset.
Truex~\etal~\cite{truex2019demystifying} further demonstrate that MI attack is data-driven and the attack models are transferable across model types by proposing various attack data generation techniques. 
Yeom~\etal~\cite{yeom2018privacy} and Salem~\etal~\cite{salem2019ml} attribute the effectiveness of MI attacks to the \textit{overfitting} of ML models. Specifically, the maximal posteriors (i.e., class confidence) of a training input tends to be higher than a non-training input. Based on this intuition, Salem~\etal~\cite{salem2019ml} propose confidence-thresholding attacks that predict highly confident samples within the training set. 
On the other hand, Yeom~\etal~\cite{yeom2018privacy} show that overfitting is a sufficient but not necessary condition for MI, and a stable and generalized model also reveals membership through black-box access. Using the average training loss as prior knowledge, they attack the target model with a threshold-based method. To increase the attack success rate when the target model is well-generalized, Long~\etal~\cite{long2020pragmatic} select valuable records from target records for MI. 
Most recently, Choquette~\etal~\cite{choquette2021label} consider a more challenging scenario, where only the predicted hard labels, rather than the confidence scores, are available. They assume that the training inputs exhibit higher robustness, thus, the empirical robustness of the predicted hard labels to the data augmentations and adversarial examples can be used to infer the membership of a record. 

In addition to the aforementioned classification models in AV, the attacker can perform MI against generative models in AV as well, as generating synthetic data for training is critical in AV training and testing. Hayes~\etal~\cite{hayes2019logan} 
used the confidence of the \textit{discriminator} over the samples to recognize the training samples, while
Liu~\etal~\cite{liu2019performing} examine the behaviors of the \textit{generator}. In particular, they trained another reconstruction network on the generator output to reproduce the target input, and 
the final reconstruction error is used to decide whether the target input is in the training set or not.

Besides MLaaS platforms, membership inference attacks can also be performed on aggregated location statistics that are commonly used in AV services.  
With prior knowledge such as users' real locations over an observation period or user participation in past aggregate groups, 
Pyrgelis~\etal~\cite{pyrgelisknock} propose a game-based procedure to infer whether a target individual's data is in the group used to calculate the shared aggregated location statistics. This membership information is helpful in mounting stronger privacy attacks such as re-constructing a user's trajectory from aggregated location statistics~\cite{xu2017trajectory}.

\underline{Model Inversion Attacks:}
Model inversion attacks focus on inferring the private \textit{statistics} or attributes of the training data for a machine learning model~\cite{Fredrikson2015ccs}. 
Fredrikson~\etal~\cite{Fredrikson2015ccs} extract an average representation of the training inputs for each class from the trained models. For example, they recover the average face for a particular person from the face recognition model via optimization-based methods (\eg, gradient descent).
However, their attack targets are simple models such as logistic regression, decision trees, and shallow neural networks in the context of face recognition. For deep neural networks, where the attack optimization is non-convex and the training data are high-dimensional, the optimization-based methods may easily stuck in local minima, and the generated features are unrealistic. 
To overcome these limitations, Yang~\etal~\cite{yang2019neural} leverage the public data to regularize the inversion, and train a separate network that swaps the input and output of the target network to perform MI attacks, which can be trained in a black-box setting. 
Besides leveraging public data, Zhang~\etal~\cite{zhang2020secretrevealer} further used generative adversarial networks (GAN)~\cite{goodfellow2020generative} to generate data that share similar distribution with private data. Utilizing GANs as well as the corrupted version of the private image as auxiliary knowledge, they recover the sensitive information in an image by solving an optimization problem.
Note that differential privacy is designed to protect the privacy of individual \textit{sample} membership, so differentially private models are still vulnerable to model inversion attacks that aim to infer sensitive \textit{attributes} of training input~\cite{bhowmick2018protection}.

Due to the broadness and complexity of AV data, model inversion attacks on AV models  pose a more significant risk by revealing sensitive locations and other personal information.

\underline{Model Memorization Attacks:}
Model memorization attacks focus on recovering the \textit{exact} feature values of the training data from a machine learning model~\cite{song2017machine}, which can be conducted in both white-box and black-box settings. 
Carlini~\etal~\cite{carlini2019secret} show that one can recover specific training records from the trained generative sequence models, such as credit card numbers. Specifically, to investigate the models' unintentional memorization of training data, they insert specific random sequences into the training data and verify that the language models memorize those sequences via testing. 
As different large language models are equipped in AV for better interaction and recommendation, such model memorization attacks have caused great privacy concerns in the AV industry.

\subsection{Protection for Individual Privacy}
\label{sec:individual-protection}
Based on the privacy risks in individual privacy, here we summarize the corresponding protection methods for individual privacy. In particular, we categorize these protection strategies into five categories: \textit{naive data anonymization, K-anonymization, differentially private learning, general private data synthesis, differentially private data synthesis, and encryption} (see Table~\ref{tb:ind_protection}).

\subsubsection{Naive Data Anonymization}\label{sec:naive_data_anony}

Naive anonymization aims to directly remove identifiers from data, such as faces, names, and addresses of the users, in order to protect individual privacy. 
For \textbf{structured} data like location, naive anonymization methods include performing geometric transformations (shifting, rotating) to obscure true position~\cite{punhani2020privacyberkeley}. Moreover, one can directly report several false positions. However, these naively anonymized locations fail to credibly imitate the mobility of living people, so their utility is limited, and they can be easily broken by inference attacks.
For \textbf{unstructured} data like images and videos, which have higher dimensions and are more complex than structured data,  existing anonymization methods usually blur visual objects (\eg, faces and license plates) in the image or video frame to preserve privacy. Such anonymization methods include obfuscation with a solid colored box, pixelization, random pixel shuffling, Gaussian blur, and distortion. For example, Silva~\etal~\cite{datoward} detect the texts in street scenarios, query Google Places API to improve accuracy, and then blur the text to preserve privacy.
On the other hand, various empirical evaluations are conducted to study the impact of data anonymization on the ML model utility. 
Schnabel~\etal~\cite{impact2019ai4i} show that the impact of anonymizing training data on the detection performance of vehicles depends on various factors, such as the type and size of objects, anonymization methods, and proportion of the object type in the dataset. 
Moreover, Krontiris~\etal~\cite{krontiris2020buckle} point  out that anonymization can have a different impact on detection performance, depending on how important the region we target is for feature learning. For example, perhaps blurring license plates can have a different impact on the performance of a car detector, compared with the impact of blurring faces on the human detector.
More advanced methods are proposed for privacy protection while maintaining a high utility level. In particular, Erdelyi~\etal~\cite{erdelyi2014adaptive} present a privacy filter based on cartoons. Experiments show that cartooning achieves visually appealing results, which maintain higher intelligibility than blurring and especially pixelation. 

Nevertheless, naive data anonymization is insufficient for data privacy in AV systems because there may be multiple observations of an individual over time. Blurring the face or license plate does little to preserve the privacy of the individual, who could be tracked across observations by their clothing or other features~\cite{KazhamiakaZB21}. Advanced strategies of data redaction, such as silhouette redaction via image segmentation~\cite{Orekondy2018}, are needed.

\subsubsection{K-Anonymization}\label{sec:k_anonymization}
K-anonymization has been widely used to anonymize the location and trajectory data. 
The original widely-used definition of $k$-anonymity~\cite{sweeney2002k} requires that the attacker cannot infer  which user is executing the query, among a set of $k$ different users. In the interest of protecting users' location, $k$-anonymity requires the location of each user to be indistinguishable among a set of $k$ users. One way to achieve this is to use \textit{dummy locations}~\cite{dummies2005}, e.g. generating $k-1$ appropriate dummy locations and using the real and dummy locations simultaneously. Another method to achieve $k$-anonymity is through \textit{cloaking}~\cite{duckham2005formal,gedik2007protecting, xu2009privacy, wang2016providing}, which involves creating a cloaking region that includes $k$ locations sharing certain properties of interest.
However, k-anonymization has been shown to achieve low performance on the anonymization of high-dimensional datasets~\cite{aggarwal2005k}.

Unfortunately, anonymized location data can be re-identified with high probability via de-anonymization algorithms~\cite{taeihagh2019governing}. For example, the de-anonymization attack proposed by~\cite{gambs2013} can infer the identity of a particular individual behind a set of mobility traces in an anonymized geo-located dataset.

\begin{table}[t]
\scriptsize
\caption{\textbf{Individual privacy: protection methods. }
}
\vspace{-2mm}
\centering
\resizebox{1\columnwidth}{!}{%
\begin{tabular}{ llllllll}
\toprule[1pt]
\textbf{Defense Category} & \textbf{Protection Target}     & \textbf{Methodology} & \textbf{Data Type} \\ 
\midrule[1pt]
\multirow{3}{*}{\makecell[l]{Naive Data \\ Anonymization}}     &\multirow{3}{*}{Raw data}    & \multirow{2}{*}{Obfuscation}   & Location~\cite{punhani2020privacyberkeley}   \\ 
&  &        & Video~\cite{datoward}     \\ \cmidrule{3-4} 
& &       Cartooning & Video~\cite{erdelyi2014adaptive}    \\  \cmidrule{1-4}
\multirow{2}{*}{K-Anonymization } & \multirow{2}{*}{Raw data}     &  Dummy location & Location~\cite{dummies2005}   \\  \cmidrule{3-4}
&    &  Cloaking  & Location~\cite{duckham2005formal,gedik2007protecting, xu2009privacy, wang2016providing}     \\ \cmidrule{1-4}
\multirow{10}{*}{\makecell[l]{Differentially \\Private\\ Learning}}   &  \multirow{4}{*}{Raw data}&    \multirow{4}{*}{Input pertb}  & Location~ \cite{andres2013geo,bordenabe2014optimal,chatzikokolakis2015location}   \\  
   & &     & trajectory~\cite{han2018research,ma2019real}   \\  
   & &      & eye-tracking~\cite{david2021privacy}     \\ \ 
  &    &   & audio~\cite{ahmed2020preech}  \\\cmidrule{2-4}
 &   \multirow{3}{*}{\makecell[l]{Supervised \\ Learning}}  &  Objective pertb  & general~\cite{chaudhuri2011differentially,kifer2012private}   \\ \cmidrule{3-4}
 &   &  Output pertb  & general~\cite{chaudhuri2011differentially, zhang2017efficient}  \\ \cmidrule{3-4}
 &   &  Gradient pertb  & general~\cite{abadi2016deep}      \\ \cmidrule{2-4} 
 &  {\makecell[l]{Semi-supervised \\ Learning}}  &   Objective pertb 
 & general~\cite{papernot2016semi, papernot2018scalable}   \\ \cmidrule{2-4}
  & \multirow{3}{*}{\makecell[l]{Reinforcement \\ Learning}}  &  Objective pertb  & sequential data~\cite{wang2019privacy}     \\ \cmidrule{3-4}
&      &  Output pertb & sequential data~\cite{balle2016differentially}   \\ \cmidrule{3-4}
&    &  Gradient pertb   & sequential data~\cite{xie2019privacy, seo2020differentially, ono2020locally}   \\ \cmidrule{1-4}
 \multirow{2}{*}{\makecell[l]{General Private  Data Synthesis}}   & \multirow{2}{*}{Generative modeling}  &  \multirow{2}{*}{\makecell[l]{GAN-based}}  & vision~\cite{krontiris2020buckle, wu2018privacy,Brkic2017, pittaluga2019learning,raval2017protecting,xiong2019privacy}   \\
&  &   & text~\cite{li2018towards}    \\ \cmidrule{1-4}
\multirow{2}{*}{\makecell[l]{ Differentially  Private\\ Data Synthesis} }  &  \multirow{2}{*}{Generative modeling}    &  DP-SGD-based & tabular, image~ \cite{xie2018differentially, torkzadehmahani2019dp, chen2020gs} \\  \cmidrule{3-4}
  &   &  PATE-based & tabular, image~\cite{yoon2018pategan, long2019scalable,wang2021datalens}  \\  \cmidrule{1-4}

\multirow{4}{*}{Encryption }        & \multirow{2}{*}{\makecell[l]{Raw data}}    & Encrypting data  & general~\cite{bost2015machine, gilad2016cryptonets}  \\ \cmidrule{3-4}
&  & MPC & general~\cite{ben1988completeness,bonawitz2017practical}  \\ \cmidrule{2-4}
& \multirow{2}{*}{\makecell[l]{Supervised \\ Learning}}     & Encrypting model  & general~\cite{aono2017privacy}     \\ \cmidrule{3-4}
&  & MPC & general~\cite{ben1988completeness,bonawitz2017practical}  \\ 
\bottomrule
\end{tabular}}
\vspace{-7mm}
  \label{tb:ind_protection}
\end{table}

\subsubsection {Differentially Private Learning}

Differentially Private (DP) can be widely applied to protect different types of AV private data with formal privacy guarantees when they are used to train machine learning models. 
We categorize the DP learning methods based on different DP mechanisms: \textit{input perturbation, objective perturbation,  gradient perturbation}~\cite{yu2019gradient}, and \textit{output perturbation}. For each of them, we summarize the general DP methods in supervised learning  and reinforcement learning (RL), which can be adapted to train differentially private models in relevant AV scenarios.

\underline{Input perturbation} refers to perturbing the input data by DP noise. It is related to Naive Data Anonymization in section ~\ref{sec:naive_data_anony}, but it requires the added noise to satisfy DP.  To achieve differential privacy in the input perturbation stage, DP noise proportional to the scale of the sensitivity of input data is added to the input data.  As long as the input perturbation step satisfies DP guarantees, the entire pipeline will be DP due to the post-processing property of DP.

DP is used to provide privacy guarantees for location-based systems against membership disclosure, where location is perturbed by DP noise before it is released ~\cite{andres2013geo,bordenabe2014optimal,chatzikokolakis2015location}. 
In terms of trajectory data, Han~\etal~\cite{han2018research} develop a spatial-division-based method to protect location and trajectory privacy with DP. To improve the utility, Ma~\etal~\cite{ma2019real} adopt a dynamic interval sampling method for differentially private real-time trajectory data release. Specifically, in the sampled time-stamp, the real position perturbed by DP noise is used, otherwise, the predicted position is used.
To protect user identification in eye-tracking technology, Gaussian noise is added to the collected gaze position~\cite{david2021privacy}. 
As for speech recognition on audio, Preech~\cite{ahmed2020preech} adds noise words to protect the acoustic features of the speakers’ voices as well as the textual content.

\underline{Objective perturbation} achieves DP by adding noise to the objection function (i.e., the empirical loss) and then solving the minimization problem w.r.t. the perturbed objective function. 
In \textbf{supervised} learning, Chaudhuri~\etal~\cite{chaudhuri2011differentially} are the first to work on differentially private empirical risk minimization (ERM). They assume the loss function is L-Lipschitz and provide the theoretical analysis of the noise bound and the excess empirical risk bound. Kifer~\etal~\cite{kifer2012private} further extend the analysis of~\cite{chaudhuri2011differentially} to convex ERM problems. Note that the above methods require strong assumptions on the objective function to bound the sensitivity. 
In \textbf{RL} models within AV for control with private rewards and public states, Wang~\etal~\cite{wang2019privacy} propose a differentially private Q-learning algorithm in a continuous space to protect the privacy of the value function approximator by adding functional noise to the value function iteratively during training. 
Notably, to control the sensitivity, they derive that the reproducing kernel Hilbert space (RKHS) norm of the noised function can be bounded.

When the model objective is more complex, such as for DNNs, advanced works have been proposed to leverage  \textbf{aggregation}, which can also reduce the required privacy budgets so as to achieve tighter privacy guarantees.
For instance, the private aggregation of teacher ensembles (PATE)~\cite{papernot2016semi, papernot2018scalable} in \textbf{semi-supervised} setting perturbs the training objective of a \textit{student} model. Specifically, it leverages the noisy aggregation of feedback from \textit{teacher} models to supervise the training of the student model, and
it is proved that predictions made by the majority of teacher models should not be affected by a single training sample, which ensures strong DP guarantees.


\underline{Gradient perturbation} aims to
first \textit{clip} the model gradient to control its sensitivity, and then \textit{perturb} the gradient by noise at {each} training step to ensure DP. 
DP-SGD~\cite{abadi2016deep} is first proposed in \textbf{supervised} learning, which trains the model by introducing noise in each update of the model parameters, thus providing DP guarantees to the training procedure. Specifically, noise is added to the gradient 
where the noise magnitude is proportional to the clipped gradient norm. 
In \textbf{RL}, GPOPE~\cite{xie2019privacy} 
adds noise to stochastic gradient descent updates, aiming to guarantee privacy for off-policy evaluation.
Seo~\etal~\cite{seo2020differentially} perform DP-SGD
during the actor’s learning process of policy gradient algorithm to guarantee DP for an actor and its eligibility trace.

\underline{Output perturbation} refers to directly adding DP noise onto the output,  where the noise is proportional to the sensitivity of the output, i.e., the maximum  change of the value regarding one input change. 
In \textbf{supervised} learning, after training the model, Chaudhuri~\etal~\cite{chaudhuri2011differentially} directly add noise to the model parameters before releasing the model. Zhang~\etal~\cite{zhang2017efficient} propose to first smooth objectives, then use gradient descent to update the model, and finally perform output perturbation. By adding noise to approximate solutions (from smooth objectives) instead of exact solutions, they achieve a better convergence guarantee for strongly convex and non-convex objectives with privacy guarantees. 
In \textbf{RL} models for control problems, \cite{balle2016differentially} propose the first private algorithm for on-policy evaluation with linear function approximation using output perturbation. Concretely, they first run an existing (non-private) least-squares policy evaluation method, resulting in a real-valued parameters vector, and then add Gaussian noise to each element of the vector.



\subsubsection{General Private Data Synthesis}
In order to de-identify subjects in unstructured data such as images or videos, while preserving non-identity-related aspects of the data and consequently enabling better AV data utility, GAN-based methods are proposed to replace faces or number plates in the video with generated ones~\cite{krontiris2020buckle}.  In \cite{wu2018privacy,Brkic2017}, generative full body and face de-identification methods are proposed to avoid the recognition of human ID or other biometric and non-biometric identifiers such as hair color, clothing, hairstyle, and personal items, while preserving data utility.
GAN-based visual secrets protection methods are later introduced by~\cite{pittaluga2019learning,raval2017protecting}, in which the authors use a generator as an obfuscator to decrease the probability of successfully detecting secret pixels. 
To protect text privacy in the video (e.g., street name, shop name), GAN-based methods are used to explicitly obscure
important characteristics at training time for text representations~\cite{li2018towards}.
The aforementioned GAN-based methods mainly focus on small objects, e.g., faces and numbers, which are easy to detect and modify. However, when it comes to location privacy in camera data, these methods cannot be applied directly because street view images have more complex context structures containing a variety of objects. With the aim to resist location-inference attack for camera data in auto-driving, Xiong~\etal~\cite{xiong2019privacy}  integrate GAN and image-to-image translation~\cite{isola2017image} to generate privacy-preserving synthesized images with a low recognition accuracy of sensitive information.

\subsubsection{Differentially Private Data Synthesis}

To provide mathematically rigorous DP guarantees for data synthesis, various DP algorithms are proposed in the general domain, while it is less studied in the AV domain. In particular, these methods can be classified into two categories: DP-SGD-based generative models and PATE-based generative models. 

\underline{DP-SGD-based Methods:}
This line of work directly adapts DP-SGD~\cite{abadi2016deep} to GAN by clipping and perturbing gradients.
For example, 
DP-GAN~\cite{xie2018differentially} perturbs the gradients of \textit{the discriminator} with Gaussian noise at each training step because the discriminator directly accesses the private data. 
DP-GAN can be applied to both structured data, such as Electronic Health Records (EHR), as well as unstructured datasets, such as MNIST. 
A similar work DP-CGAN~\cite{torkzadehmahani2019dp} additionally generates labels using Conditional GAN~\cite{mirza2014conditional}.
GS-WGAN~\cite{chen2020gs} assumes that discriminator is not accessible by attacker thus focusing on DP training of \textit{generator}. To improve data utility, GS-WGAN uses Wasserstein loss~\cite{arjovsky2017wasserstein}, which enables the optimal clipping threshold for the gradients of the generator based on the theoretical property of Wasserstein GANs. 

\underline{PATE-based Methods:}
PATE-GAN~\cite{yoon2018pategan} adapts the PATE framework~\cite{papernot2016semi, papernot2018scalable} by training multiple teacher discriminators and uses them to update the student discriminator, which achieves high performance on low-dimensional tabular datasets. 
G-PATE~\cite{long2019scalable} improves upon PATE-GAN by directly training a student generator using the teacher discriminators. Moreover, it uses random projection to reduce the gradient dimension during training, and thus can be applied to both structured data and unstructured image data. 
Most recently, DataLens~\cite{wang2021datalens} 
proposes  a DP gradient compression and aggregation approach, which combines top-$k$ dimension compression with a corresponding noise injection mechanism, 
and demonstrates utility improvement on high dimensional datasets empirically.


 

\subsubsection{Cryptography}
This line of literature is based on standard cryptography concepts like secret sharing~\cite{beimel2011secret}, private information retrieval~\cite{chor1995private} and symmetric encryption techniques~\cite{bellare1997concrete}.
Encryption methods can be divided into two groups:  \textit{encrypting training data} and \textit{encrypting ML model}.
\textbf{(1)} In terms of encrypting training data, Bost \etal~\cite{bost2015machine} express the training algorithm as a low-degree polynomial and train over encrypted data in three classifiers: hyperplane decision, Naive Bayes, and decision trees. CryptoNets~\cite{gilad2016cryptonets} converts the learned neural networks to make them applicable to encrypted input data.
\textbf{(2)} In terms of encrypting ML model,
Aono~\etal~\cite{aono2017privacy} use additively homomorphic encryption on the gradients during the collaborative learning, which can prevent information leakage to the honest-but-curious server. 
Beyond the above methods, secure multi-party computation (SMC) is the extension of encryption under the multiparty setting ~\cite{ben1988completeness,bonawitz2017practical} to ensure both data and model privacy. 





\textbf{Discussion.}
In order to protect individual privacy, various privacy-preserving methods  have been proposed for structured and unstructured data. 
Naive data anonymization, k-anonymization, and general private data synthesis are common protection methods, but they can be easily broken without formal guarantees. 
Differential private learning and differentially private data synthesis provide strong privacy guarantees. They are widely applied for structured data like location and trajectory, but they are not well developed for unstructured AV data. This is because in AV, the goal is not to protect \textit{individual data sample}, but the \textit{individual users}. In order to achieve that, one may first detect the individual user in images, videos,  point clouds, etc., and then make the user differentially private. 
Cryptography methods provide strict privacy protection, but they usually incur high computation costs.

Another line of work suggests using regulations 
to restrict the AV company and AV users from collecting private user data as well as public data in the street from the legal viewpoint, which are orthogonal to the concrete privacy-preserving techniques and serve as supplements.  
Taeihagh~\etal ~\cite{taeihagh2019governing} suggest increasing transparency regarding what AV data is being collected, who is using it, how the data is being used and shared, in-depth disclosures about potential security vulnerabilities, providing opt-out options, and limiting data collection to the minimal amount required.
We refer the readers to  \cite{MULDER2021105530,taeihagh2019governing,ioannis2020cscs} for the detailed summary of AV-related privacy regulations in the US, Europe, China, etc.

\section{Population Privacy in Autonomous Vehicles}
\label{section:population_privacy}
\textbf{Overview.}
Different from other ML applications, data collected in AV contains privacy sensitive information about not only   \emph{individuals} but also  \emph{population} level properties. For example, aggregated location statistics can reveal properties of the city, such as the location of important infrastructures and popular destinations~\cite{glancy2012privacy}.
On the one hand, population-level sensitive information is contributed by multiple individuals (i.e., multiple vehicles or AV users), but does not only depend on individual privacy. 
Consequently, removing/hiding partial individual's data from the database cannot protect the population privacy. On the other hand, 
the latent properties of a population which may represent \textit{exactly the desired utility} of the data collection, when learned, can in turn compromise the privacy of an individual~\cite{sanches2014knowing}.

In this paper, we use \emph{population privacy} to characterize the leakage of sensitive population-level information derived from AV data contributed by more than one individual. Formally, suppose a dataset $x=\{x_1, x_2, \dots, x_n\} $ is drawn from distribution $\mathcal{D}$, population privacy studies the sensitive information learned from function $f(x)$. 
Although there is a lack of discussion about population privacy in prior work, model inversion attacks against ML models have raised concerns about leaking sensitive population-level information. Moreover, in AV, there is a higher risk of leaking sensitive population-level information due to the extensive amount of data collected by external cameras about the environment~\cite{bloom2017self} and the rich information contained in vehicle location traces. 


\begin{table}[t]
\scriptsize
\caption{\textbf{Population Privacy: Risks and Protection Methods.}}
\vspace{-2mm}
\centering
\resizebox{1\columnwidth}{!}{%
\begin{tabular}{ llllllll}
\toprule[1pt]
\multicolumn{1}{c}{\textbf{Scenario}} & \textbf{Attack Target}     & \textbf{Protection Method}   \\ 
\midrule[1pt]
\multirow{2}{*}{\makecell[l]{V2X \\ communications}}        & identity~\cite{huang2020}  &  pseudonyms, encryption~\cite{huang2020} \\  \cmidrule{2-3}
& location~\cite{huang2020}   &  pseudonyms, obfuscation, caching~\cite{huang2020}    \\  \cmidrule{1-3}
\multirow{2}{*}{\makecell[l]{Federated \\ Learning in AV}}        & reconstruction~\cite{NEURIPS2019_deepleakage,huang2021evaluating,geiping2020inverting,bhowmick2018protection}   &  instance-level DP perturbation~\cite{malekzadeh2021dopamine} \\   
& population properties~\cite{melis2019exploiting}    &   user-level DP perturbation~\cite{geyer2017differentially,mcmahan2018learning,agarwal2018cpsgd} \\  \cmidrule{1-3}
\multirow{2}{*}{\makecell[l]{AV \\ services}}        & location~\cite{huang2020}  & \makecell[l]{perturbation~\cite{andres2013geo,bordenabe2014optimal,chatzikokolakis2015location},  \\ k-anonymity~\cite{gedik2007protecting, xu2009privacy, wang2016providing,sweeney2002k, wang2019achieving}} \\  \cmidrule{2-3}
& \makecell[l]{location, trip time \\ and route ~\cite{sherif2016privacyridesharing, hadian2019privacytimesharing,valetparking19}}  & encryption~\cite{sherif2016privacyridesharing, hadian2019privacytimesharing,valetparking19}  \\  \cmidrule{1-3}
AV sensor networks & \makecell[l]{mass surveillance~\cite{glancy2012privacy,calavia2012semantic, thomas2015combining} \\  road conditions~\cite{garg2018ubiquitous}}  & / \\
\bottomrule
\end{tabular}}
\vspace{-7mm}
  \label{tb:ppl_atk_pro}
\end{table}

\subsection{Privacy Risks on Population-Level}
Here we will introduce the potential privacy problems on
population-level, including private information leakage from \textit{V2X communication, federated learning in AV, AV services}, and \textit{AV sensor networks}, which are summarized in Table~\ref{tb:ppl_atk_pro}.

\subsubsection{V2X communications}
By vehicle-to-vehicle (V2V) and roadside vehicle-to-infrastructure  (V2I) communications, an attacker may eavesdrop on private sensitive information of drivers and other AV users on different levels.  For example, these communications expose the  geographical location and movements of vehicles to external networks~\cite{taeihagh2019governing}.
Formally, the privacy issues in V2X vehicular networks include identity privacy and location privacy where identity privacy refers to the information about ``who'' sends the message, and location privacy refers to the information about ``where'' the sender is~\cite{huang2020}. While identity privacy can be characterized under individual privacy, location privacy will reveal population-level sensitive information.

\subsubsection{Federated Learning in AV}
Federated Learning (FL)~\cite{mcmahan2017communication} has been widely applied to trained shared models across AVs~\cite{posner2021federated}, so that raw data can be kept in local vehicles/devices without sharing. Elbir~\etal~\cite{elbir2020federated} present an FL-based framework to train ML models 
for vehicular networks, and enlist several applications such as autonomous driving, infotainment, and route planning.
Zeng~\etal~\cite{zeng2021federated}  leverage FL for collaborative training of autonomous controller model across a group of connected AVs.
However, based on the FL privacy literature, it is possible for attackers to 
\textit{reconstruct} the training data
~\cite{NEURIPS2019_deepleakage,huang2021evaluating,geiping2020inverting,bhowmick2018protection} or infer the population \textit{properties} about training data~\cite{melis2019exploiting}  from the communicated model updates or gradients. Thus, both traditional privacy concerns in FL and population-level information leakage in FL within AV systems would raise  privacy concerns.

\subsubsection{AV Services}
Autonomous vehicles benefit from cooperation with the industrial internet and cyber-physical systems that provide AV services. In general, AV services are query-based, where users  submit  service queries to the service provider such as the destination, and these requests are tagged with their current locations to achieve spatial-temporal-related services~\cite{wang2019achieving} such as car towing. 
In each specific AV service, different types of sensitive information can be leaked. For example, in the ride-sharing scenario where the organization requires the users to disclose sensitive detailed information not only on the pick-up/drop-off locations but also on the trip time and route~\cite{sherif2016privacyridesharing} for planning purposes. 
In a time-sharing scenario, AV owners share their vehicles with others at their unwanted times, which requires the disclosure of users’ locations and route information~\cite{hadian2019privacytimesharing}. Such information can be used to infer population-level information, such as traffic patterns and the location of popular destinations.

\subsubsection{AV Sensor Networks}
In addition to the autonomous vehicles themselves, the sensors such as cameras equipped on AVs can form a large mobile sensor network, and the data aggregated from these sensors will provide a detailed current view of the physical world~\cite{garg2018ubiquitous,calavia2012semantic, thomas2015combining,glancy2012privacy}. In particular, Glancy~\etal
~\cite{glancy2012privacy} elaborate on the \textit{targeted} surveillance of a particular person and \textit{mass} (i.e., indiscriminate and comprehensive)  surveillance of groups or populations using AV. 
Recent works have also shown that video surveillance systems can be built by combining the data collected by multiple cameras in a dense camera network~\cite{calavia2012semantic, thomas2015combining}.
Garg~\etal~\cite{garg2018ubiquitous} show that ubiquitous sensing systems can be used to evaluate road conditions~\cite{garg2018ubiquitous}, and the same technique can be extended to infer sensitive information about the overall physical environment. 

\subsection{Protection for Population Privacy}
Based on the potential privacy risks at the population level above, we will next characterize corresponding protection strategies and challenges, which are summarized in Table~\ref{tb:ppl_atk_pro}.
\subsubsection{Privacy-Preserving Communications}
Existing works 
apply encryption, pseudonyms, or obfuscation to protect data in communication. Specifically, identity privacy can be protected by replacing the identifiers with pseudonyms or using group private keys to sign messages anonymously~\cite{huang2020}. The protection of location privacy includes changing pseudonyms, obfuscating vehicle locations, and reducing the number of requests sent from vehicles to the server via caching~\cite{huang2020}. 

\subsubsection{Privacy-Preserving Federated Learning}
Existing works in FL mainly concern two types of privacy, namely, \textit{user-level privacy} and \textit{instance-level privacy}~\cite{mcmahan2018learning}. In particular, in differential private FL (DPFL), the two privacy notions imply that  the trained FL model should not differ much if one user (user-level) or one instance is modified (instance-level privacy), respectively. Currently, no protection methods are specifically designed for population privacy in FL under AV settings, thus, we will briefly introduce the prior works on user-level DP and instance-level DP for standard FL models, which may alleviate the risks of population privacy to some extent.
In user-level DPFL, the information sent from the clients to the server is usually perturbed to ensure DP. Specifically, the server clips the norm of each local update, and adds Gaussian noise on the summed update~\cite{geyer2017differentially,mcmahan2018learning}. Notably, user-level DPFL has been applied for large-scale  language models with millions of users~\cite{mcmahan2018learning}, and a small user selection ratio at each FL round is one key to saving privacy budget based on the privacy amplification by subsampling~\cite{balle2018privacy}. In CpSGD ~\cite{agarwal2018cpsgd}, each user clips and quantizes the model update, and adds noise drawn from Binomial distribution, achieving both communication efficiency and DP.
Instance-level DPFL is a more direct extension of the standard DP. Basically, each user trains the local model with DP-SGD to ensure DP~\cite{malekzadeh2021dopamine}. 

\subsubsection{Privacy-Preserving AV Services}
For query-based AV services, the privacy of sensitive query contents such as location information can be preserved by noise perturbation~\cite{andres2013geo,bordenabe2014optimal,chatzikokolakis2015location}, where the perturbed location is released to the service provider.
Other strategies are $k$-anonymity~\cite{sweeney2002k} which hides the actual query content among $k$ queries, or cloaking~\cite{gedik2007protecting, xu2009privacy, wang2016providing} which hides the user among $k$ users, as discussed in Section~\ref{sec:k_anonymization}.
To improve the protected data utility, Wang~\etal~\cite{wang2019achieving} propose personalized $k$-anonymity, allowing users to specify the minimum anonymity level for each query. 
In each specific AV service, tailored privacy-preserving algorithms are proposed to protect corresponding information. For example, 
in the ride-sharing services, in order to address the unique privacy risks of pickup/drop-off locations, trip time, and route, Sherif~\etal~\cite{sherif2016privacyridesharing} present a similarity measurement technique over encrypted data for service providers, thus 
achieving privacy-preserving ride arrangements.
In the time-sharing services, in order to protect users’ locations and route information, Hadian~\etal~\cite{hadian2019privacytimesharing} design a matching scheme that optimizes the available AV for users and then propose a privacy-preserving scheduling scheme based on encryption for assigning users to AVs without sharing their exact locations and route details.
In automated valet parking (AVP) services, Ni~\etal~\cite{valetparking19} proposed a framework based on encryption to achieve secure authentication for vehicle remote pickup and user privacy protection.

\textbf{Discussion.}
By connecting to vehicle networks or cloud (e.g., V2X communication, FL, AV services, sensor networks),  the driving performance and riding experience are improved via collaboration, whereas such processes expose sensitive information.
Existing protections on V2X communications, FL, and AV services mainly perform  obfuscating, DP perturbation or encryption to protect sensitive information flow in AV networks. 
However, obfuscating and DP cannot provide guarantees for population privacy  while encryption comes with high computation cost, thus, a formal \textit{population privacy} definition with efficient algorithms and rigorous privacy analysis requires further investigation.
It is also not clear how individual and population privacy correlate under realistic privacy attacks, and to what extent 
the protection techniques in individual privacy (See Section~\ref{sec:individual-protection}) may 
alleviate population privacy.

In population privacy, the sensitive information that we want to protect can be closely correlated with the information  needed by AV services. For example, aggregated location information is important for route planning, but it can also leak the location of sensitive infrastructures.
How to balance the trade-off between utility and privacy is still an open problem for future research.

\section{Proprietary Privacy in Autonomous Vehicles}
\label{section:proprietary_privacy}
\textbf{Overview.}
Both individual and population privacy are about the sensitive information \emph{external} to an autonomous vehicle system, such as users, pedestrians, and the environment. Meanwhile, the information about the \emph{internal} system of an AV is also valuable to the companies that design, develop, and manufacture the vehicle.  
Specifically, the research and development of ML models in AV are one of the most costly investments for most AV companies. Therefore, the algorithms, model structures, and parameters are often considered valuable assets to the model owners, and the leakage of this information is considered a huge privacy threat. In this section, we use \emph{proprietary privacy} to categorize the privacy risk related to the proprietary information about AV systems. Below, we formalize the definition of proprietary privacy of machine learning models.
\begin{definition} [Proprietary Privacy]
Let $\mathcal{F}$ be the target model space, $\Theta$ be the space of sensitive parameters, and $\mathcal{A}: O \mapsto \Theta$ be an attack that infers the sensitive parameters through an oracle $\mathcal{O}: \mathcal{F} \mapsto O$. $\mathcal{F}$ is $(\varepsilon, \delta)$-\emph{proprietary private} against attack $\mathcal{A}$ through oracle $\mathcal{O}$ if 
$\Pr_{f_\theta \in \mathcal{F}}[d(f_\theta, f_{\mathcal{A}(\mathcal{O}(f_\theta))})<\varepsilon] > 1 - \delta$,
where $d$ is the metric for the difference between the target model and the reconstructed model, which is determined by the objective of the attack on proprietary privacy.
\end{definition}

The attacks on proprietary privacy are called \textit{model extraction} attacks. In this section, we formalize the definition of model extraction attacks, summarize the attack methods and the protection methods (see Table~\ref{tab:model_extraction}), and discuss the threats and research opportunities related to proprietary privacy. 



\subsection{Privacy Risks on Proprietary-Level}
\subsubsection{Proprietary Privacy Attack Definition and Taxonomy}
The \textit{model extraction attacks}, also known as model stealing attacks, aim to infer sensitive information about ML models through model \textit{prediction APIs} or \textit{side channels}. 
We formalize the definition of model extraction attacks proposed in~\cite{tramer2016stealing}: 
\begin{definition}[Model Extraction Attack $\mathcal{A}_{\rm ME}$]
Let $f_\theta \in \mathcal{F}$ be the target ML model and $\mathcal{O}: \mathcal{F} \mapsto O$ be an oracle that reveals information about $f_\theta$. A model extraction attack $\mathcal{A}_{\rm ME}( \mathcal{O}(f_\theta))$ outputs the reconstructed parameters $\theta'$ so that the difference between the $f_\theta$ and $f_{\theta'}$ represented by $d(f_\theta, f_{\theta'})$ is small. 
\end{definition}
\textit{First,} the parameter $\theta$ defines the \textit{sensitive information} a model extraction attack aims to infer. For example, in the model parameter stealing attack~\cite{tramer2016stealing}, the attacker knows the architecture of the target model $f_\theta$ and aims to infer the parameters of the model. Besides model parameters, model extraction attacks can also target model architecture (e.g., layer depth and dimensions) or hyperparameters (e.g., learning rate and regularization factors)~\cite{hu2020deepsniffer}. 
\textit{Second,} the oracle $\mathcal{O}$ characterizes the \textit{attack channels} and the attacker's \textit{additional knowledge} about $f$. Based on the attack channels, model extraction attacks can be divided into two main categories: attacks through \textit{prediction APIs} or \textit{side channels}. In prediction API attacks, the attackers interact with the target model by submitting queries $\mathbf{x}$ to the oracle $\mathcal{O}$. The oracle $\mathcal{O}(f_\theta; \mathbf{x})$ returns prediction labels, prediction confidence, or other intermediate results of the model on the input $\mathbf{x}$. In side-channel attacks, the oracle $\mathcal{O}$ returns side-channel information such as prediction time, electromagnetic emanations, or memory access information. 
\textit{Third,} based on the attack objective, model extraction attacks can be taxonomized into \emph{task accuracy extraction} and \emph{model fidelity extraction}~\cite{jagielski2020high}. Task accuracy extraction focuses on reconstructing models with good performance on the underlying learning task, while model fidelity extraction aims to build models that match the victim model on \textit{any} input, even for incorrect predictions. 
Model fidelity extraction 
also makes other ML attacks, such as model inversion and membership inferences easier. 
We formalize the definition of task accuracy difference $d_{\rm acc}$ and model fidelity difference $d_{\rm fid}$ in Appendix~\ref{sec:atk_obj_model_extract}, which are also discussed in~\cite{jagielski2020high}.


\begin{table}[t]
\caption{\textbf{
Proprietary privacy: risks and protection methods.
}}
\vspace{-3mm}
\centering
\resizebox{1.02\columnwidth}{!}{
\begin{tabular}{lllllll}
\toprule
{\textbf{Channel}} & {\textbf{Sensitive Info}} & {\textbf{Attack Goal}} & {\textbf{Model Type}} & {\textbf{Adversarial Knowledge}} & \multicolumn{2}{c}{\textbf{Protection Methods}} \\ 
\midrule
& & & & &  Prevention & Detection \\
\cmidrule{6-7}
\multirow{8}{*}{\makecell[l]{Prediction \\ API}} & \multirow{7}{*}{Parameters}  & \multirow{3}{*}{Accuracy} & LM, DT, NN & Input saliency map~\cite{milli2019model} & \multirow{8}{*}{\makecell[l]{Limit returned\\ information\\\cite{tramer2016stealing, lee2018defending, wang2018stealing};\\ Ensemble \\ of models\\\cite{tramer2016stealing,alabdulmohsin2014adding, alabdulmohsin2014adding}}} & \multirow{8}{*}{\makecell[l]{Abnormal\\ query\\ patterns\\\cite{juuti2019prada, kesarwani2018model}}}  \\
 &  &  & LM, DT, NN & Prediction labels~\cite{tramer2016stealing, chandrasekaran2020exploring} & & \\ 
 &  &  & BERT & \makecell[l]{Prediction confidence\\ or answer spans~\cite{krishna2019thieves}} & & \\ \cmidrule{3-5}
 &  & \multirow{4}{*}{Fidelity} & LM & Prediction labels~\cite{jagielski2020high} & & \\ 
&  &  & ReLU NN & Prediction labels~\cite{carlini2020cryptanalytic},  \cite{rolnick2020reverse} & &\\ 
 &  &  & GNN & \makecell[l]{Node embedding or \\prediction confidence~\cite{shen2022model}} & &\\
 &  &  & RL & State action sequence~\cite{chen2021stealing} & & \\ \cmidrule{2-5}
 & Hyperparameters & Fidelity & LM, NN & \makecell[l]{Training dataset and \\model parameters~ \cite{wang2018stealing}} & & \\ \midrule
 & & & & &  Isolation & Randomization \\
\cmidrule{6-7}
\multirow{7}{*}{\makecell[l]{Side \\ Channel}} & \multirow{3}{*}{Architecture} & \multirow{3}{*}{Fidelity} & \multirow{3}{*}{NN} & \makecell[l]{Colocation and shared \\ ML library code~\cite{yan2020cache,hong2019how}} & \multirow{7}{*}{\makecell[l]{Spatial\\ isolation\\\cite{Kiriansky2018dawg, Domnitser2012non, kim2012stealthmem, liu2016catalyst};\\Temporal\\ isolation\\\cite{ferraiuolo2016lattice, Sprabery2018scheduling}}} & \multirow{7}{*}{\makecell[l]{Memory\\ address\\ randomization\\\cite{wang2007new, qureshi2018ceaser, qureshi2019new};\\Clock\\ randomization\\\cite{martin2012timewarp, trilla2018cache}}} \\
 &  &  &  & Colocation~\cite{liu2020ganred} & & \\  
 &  &  &  & EM or bus snooping~\cite{hu2020deepsniffer} & & \\ \cmidrule{2-5}
 & \makecell[l]{Architecture \& \\parameters} & Fidelity & MLP, CNN & EM and timing~\cite{batina2019csi} & & \\ \cmidrule{2-5}
 & \multirow{3}{*}{\makecell[l]{Architecture \& \\parameters}} & \multirow{3}{*}{Accuracy} & NN & \makecell[l]{Prediction confidence \\ and timing~\cite{duddu2018stealing}} & & \\ 
 &  &  & BNN & \makecell[l]{Prediction confidence \\and EM~\cite{yu2020deepem}} & & \\  
 &  &  & CNN & Off-chip memory access~\cite{hua2018reverse} & & \\ 
\bottomrule
\multicolumn{7}{c}{\makecell[c]{LM: Linear Model; DT: Decision Tree; NN: Neural Network; GNN: Graph Neural Network; \\ MLP: Multi-Layer Perceptron; CNN: Convolutional Neural Network; BNN: Binary Neural Network}} \\
\end{tabular}}
    \label{tab:model_extraction}
    \vspace{-7mm}
\end{table}
\subsubsection{Model Extraction Attacks through Prediction APIs}
The attacker can interact with the target model's prediction API by providing an input query $\mathbf{x}$. 
The oracle $\mathcal{O}(f_\theta; \mathbf{x})$ 
returns the target model's prediction on that query. 
This process can repeat multiple iterations, allowing the attacker to accumulate information about the target model. Most attacks through prediction APIs target a classification model 
$f_\theta$ 
and often assume that the attacker obtains either the model's prediction confidence vector over all the possible class labels (i.e., $\mathcal{O}(f_\theta; \mathbf{x})=f_\theta(\mathbf{x)}$) or the predicted class
(i.e., $\mathcal{O}(f_\theta; \mathbf{x})=\arg\max f_\theta(\mathbf{x)}$)~\cite{tramer2016stealing, chandrasekaran2020exploring, jagielski2020high, carlini2020cryptanalytic, rolnick2020reverse}. In addition, Milli~\etal~\cite{milli2019model} proposed model extraction attacks on model explanations by assuming that the oracle $\mathcal{O}$ returns the model's saliency map of the input query, and they demonstrate that this assumption significantly reduces the cost of model extraction attacks. Their findings highlight the trade-off between improving model explainability and proprietary privacy protection. 

Besides classification models, other models such as graph neural networks (GNN)~\cite{scarselli2008graph}, natural language models, and deep reinforcement learning (DRL) are also vulnerable to model extraction attacks. 
Shen~\etal~\cite{shen2022model} proposed 6 model extraction attacks on GNN with varying assumptions on the adversarial knowledge on the query graph $\mathbf{x}$ (i.e., whether it shares the same distribution as the training graph), and the responses of the oracle $\mathcal{O}(f_\theta; \mathbf{x})$ (i.e., the node embedding, prediction confidence vector or t-SNE~\cite{van2008visualizing} projection of prediction confidence vector). 
Their experiment results demonstrate an increasing difficulty in model extraction attacks as the information returned by $\mathcal{O}(f_\theta; \mathbf{x})$ reduces. 
Krishna~\etal~\cite{krishna2019thieves} designed model extraction attacks against BERT-based APIs~\cite{devlin2018bert} for both classification tasks and question answering (QA) tasks. 
Their attacks demonstrate that BERT models can be extracted with high accuracy with randomly sampled word sequence queries. 
Chen~\etal~\cite{chen2021stealing} extended the attacks to deep reinforcement learning, where the oracle $\mathcal{O}(f_\theta; \mathbf{x})$ returns a state action sequence. They assume that the attacker can operate the model in a normal environment, and therefore cannot manipulate states during the attack. 

Besides model parameters, model hyperparameters can also be inferred from model extraction attacks.  Wang~\etal~\cite{wang2018stealing} extract the model hyper-parameters (i.e., regularization coefficients) based on model parameters and training data. 

\looseness=-1
\subsubsection{Model Extraction Attacks through Side Channels}
The attacker infers the target model's sensitive information through collected indirect information about the system or hardware. In a recent survey of side channel attacks on neural networks~\cite{chabanne2021side}, the side-channel model extraction attacks are divided into four categories based on  attack approaches: cache attacks~\cite{yan2020cache, hong2019how, liu2020ganred}, physical access attacks~\cite{yu2020deepem, batina2019csi}, remote timing attacks~\cite{duddu2018stealing}, and memory access pattern attacks~\cite{hua2018reverse}.
\textit{First}, in a \underline{cache side channel} attack, the attacker runs a process that is co-located with the victim process and tries to determine whether a certain function has been called by the victim process. To achieve this, the attacker infers whether a target address is stored in the shared cache based on the access time of that address. Traditional cache side channel attacks often focus on extracting the secret keys used for encryption. However, recent works~\cite{yan2020cache, hong2019how, liu2020ganred} show that these attacks are capable of revealing the sensitive architectural parameters of neural networks by monitoring the number of executions for each functions such as matrix multiplication, pooling, and activation functions. Furthermore, Li~\etal~\cite{liu2020ganred} show that the attacker can infer the architecture of the target model without a shared ML library between the victim and attack process.
\textit{Second}, the \underline{physical access} attacks use side-channel signals that can be directly measured given physical access to the device running the target model. For example, DeepSniffer~\cite{hu2020deepsniffer} extracts the volume of memory reads/writes through electromagnetic (EM) emanations or bus snooping attacks, and uses these signals to infer the architecture of the target model. CSI NN~\cite{batina2019csi} combines timing and EM side channels to infer both the model architecture and the model parameters. DeepEM~\cite{yu2020deepem} focuses on binary neural networks (BNN). It measures electromagnetic emanations in addition to timing and power to infer the network architecture, and uses the model predictions to reconstruct the model parameters. 
\textit{Third}, in the \underline{remote timing} attack~\cite{duddu2018stealing}, the attacker has access to the prediction API of the target model and an accurate timing of the inference time of the model. The timing side channel is used to determine the depth of the target model, and the model parameters are reconstructed through the prediction results using reinforcement learning. 
\textit{Fourth}, Hua~\etal~\cite{hua2018reverse} proposed a \underline{memory access pattern} attack against the CNN secure accelerator. The attack exploits off-chip memory access patterns to infer structure and model parameters, and demonstrates that side-channel model extraction attacks are possible even with data encryption and secure accelerators. 

\subsection{Privacy Protection on Proprietary Level}
Next, we summarize the corresponding defenses against prediction API attacks and side channel attacks (see Table~\ref{tab:model_extraction}).
\subsubsection{Protection against Prediction API Attacks}
The defense can be divided into two categories: attack prevention and attack detection. Attack prevention focuses on designing models that are more robust to model extraction attacks.
The prevention techniques include two approaches: 
i)
limiting the amount of information returned per query by adding noise to the prediction~\cite{tramer2016stealing, lee2018defending}, rounding the predicted probabilities~\cite{tramer2016stealing, wang2018stealing}, or only returning the class output~\cite{tramer2016stealing};
ii) using an ensemble of models by either combining the prediction of the ensemble of models~\cite{tramer2016stealing} or randomly sampling from a distribution of model parameters~\cite{alabdulmohsin2014adding}. 
Besides attack prevention, model owners can also detect the attack through its abnormal query patterns~\cite{juuti2019prada, kesarwani2018model} and stop the attack before the attacker gets sufficient information to infer the model parameters.



\subsubsection{Protection against Side Channel Attacks}

The protection 
includes isolation and randomization. Partition techniques~\cite{Kiriansky2018dawg, Domnitser2012non, kim2012stealthmem, liu2016catalyst} create spatial isolation between the attack program and the target program, and scheduling techniques can create temporal isolation~\cite{ferraiuolo2016lattice, Sprabery2018scheduling} to prevent the attacks. Randomization includes memory address randomization~\cite{wang2007new, qureshi2018ceaser, qureshi2019new} and clock randomization~\cite{martin2012timewarp, trilla2018cache}, which prevent the attack program from obtaining the correct memory address or precise timing information of the target program to infer the correct cache state,  respectively.
In addition, Oblivious RAM hides the memory access pattern through data address encryption~\cite{liu2015ghostrider, liu2013memory, stefanov2018path} and can protect information on the bus.

\textbf{Discussion.}
Proprietary privacy concerns with
leaking sensitive information about the internal system of AVs, 
including ML model architectures, model parameters, and model hyper-parameters. The variety of sensitive information increases the challenge of detecting and defending against various types of model extraction attacks. Although there is plenty of research on the protection against model extraction and side-channel attacks, each protection method is often designed for specific attacks. The systematic quantification and protection of proprietary privacy remain an open and challenging problem.

\section{Insights, Challenges, and Future Directions}

\label{section:discussion_future_directions}
In this section, we summarize the characteristics and fundamental connections among different levels of privacy concerns in AV, then discuss the main challenges of corresponding privacy protection, as well as potential future directions.

\textbf{Privacy Concerns and Protection in the Life Cycle of AV.}
Privacy concerns exist throughout the entire life cycle of AV, which involves different types of data collection and protection. In this work, we provide a novel taxonomy for privacy in AV to categorize it as the \textit{individual}, \textit{population}, and \textit{proprietary} levels based on both internal and external environment data.
We show that different from traditional privacy problems for a single model or database, which mainly focuses on individual-level privacy, the privacy concerns in AV are more complex given its nature as a dynamic system.
Thus, protecting privacy in AV systems involves several new challenges:
\underline{(1)} The AV system is a combination of different types of machine learning models and pipelines, including the standard perception (e.g., object detector), prediction, control, and planning (e.g., different RL algorithms) components, as well as the functional ML models such as the eye-tracking safety enhancement model and audio/NLP based recommendation models. These models take different modalities of data (e.g., 2D images, 3D LiDAR, and text) as input and process them with diverse architectures. Thus, it is challenging to design one privacy protection scheme for all, and efficient hybrid privacy protection strategies and effective aggregation protocols are required.
\underline{(2)} Given the fact that different types of data are collected for AV training and testing, how to define the privacy requirements for these data is also challenging. For instance, the face of pedestrians may be more sensitive than their location at a specific time.
\underline{(3)} In autonomous driving scenarios, the collaborations between vehicles are important and how to ensure private communication in the dynamic V2X network poses another new challenge compared with existing privacy protection for static models.  

\textbf{Future Directions.}
Given the unique properties and challenges of privacy in AV,  we will briefly discuss some potential future directions.
\underline{(1)} Given the complex dynamic system components in AV, relying on only one type of privacy protection approach is less effective. For instance, if we ensure DP for each component, the added noise will be very large and significantly hurt the final performance. Thus, organic hybrid strategies are important to help protect privacy with the target. Specifically, it is possible to design cryptography approaches for key components with lower dimensional input and design DP and other strategies to deal with high-dimensional data protection. 
\underline{(2)} As training AV requires a large amount of data, it is very promising to generate synthetic/simulational data in a private way to protect the privacy of the original data as well as improve the training effectiveness. Thus, different privacy-preserving data generation approaches, especially those that could preserve the generated data realism, would make a great impact on protecting privacy in AV.
\underline{(3)} Due to the trade-off between privacy protection and data mining to improve performance, it is also important to provide a valuation for data
so that we are able to perform data debugging and select the most ``contributional" data to add noise or redact. This is critical considering the diverse modalities and types of data required in AV training. 
\underline{(4)} Similar to the data valuation, in the end-to-end AV systems, the component level valuation and debugging is also important to guide the component-driven privacy protection to save privacy budget and the hybrid approach design. 
\underline{(5)} In practice, the privacy guarantees of AV systems are critical. For instance, DP is one possible guarantee which is  expensive in terms of system performance tradeoff and is insufficient to address all privacy concerns such as population privacy. Other types of privacy guarantees, such as mutual information-based and population-based measurements are important to help provide practical privacy guarantees.
\underline{(6)} In the V2X settings, how to leverage graph model analysis such as designing privacy-preserving message passing algorithms and minimal communication topology to ensure effective communication is important and leads to a rich set of research problems.
\underline{(7)}
As safety is of great importance for AV, how to balance between privacy-preserving training and urgent safety-critical decision in the end-to-end AV systems requires careful system and algorithm joint design.
\underline{(8)} Traditionally, AV designers seldom consider risks on privacy. Based on this review, we believe people should take privacy as a critical factor in data collection and exploitation throughout the whole life cycle of AVs, and study the performances of different design pipelines, e.g. rules-based vs learning, with the privacy-preserving requirements.
\underline{(9)} How the privacy risks of AVs would affect equity and social good is also important to explore. \underline{(10)} The privacy of AVs requires expertise across privacy, ML, automotive, policies, and transportation. How to build up proper taxonomy and training pedagogies for people with diverse backgrounds and requirements is also a promising research direction.

\newpage
{
\scriptsize
{\linespread{0.85}\selectfont\bibliography{reference}}
\bibliographystyle{ieeetr}
}
\newpage
\appendix
\subsection{Attack objectives in model extraction attacks}\label{sec:atk_obj_model_extract}

\begin{definition}[Task Accuracy Difference $d_{\rm acc}$]
Let $\mathcal{D}_{A}$ over $\mathcal{X} \times \mathcal{Y}$ be the feature and true label distribution of the underlying learning task $\mathcal{T}$. The \emph{task accuracy difference} $d_{\rm acc}$ measures the difference in the accuracy of the two models over the learning task $\mathcal{T}$. That is,
\begin{equation*}
d_{\rm acc}(f_\theta, f_{\theta'}) = \min({\rm Acc}(f_\theta; \mathcal{D}_{A}) - {\rm Acc}(f_{\theta'}; \mathcal{D}_{A}),0), 
\end{equation*}
where ${\rm Acc}(f; \mathcal{D}_{A}) = \Pr_{(\mathbf{x},y)\in \mathcal{D}_{A}}[\arg\max(f(\mathbf{x}))=y]$.
\end{definition}
The goal of task accuracy attacks is to match (or exceed) the accuracy of the target model $f_{\theta}$ on the underlying learning task, and the attackers don't need to reproduce the mistakes in $f_{\theta}$ or to mimic the predictions $f_{\theta}$ outside $\mathcal{D}_{A}$. 
\begin{definition}[Model Fidelity Difference $d_{\rm fid}$]
Let $\mathcal{D}_{F}$ over $\mathcal{X}$ be the feature distribution of the target model. The \emph{model fidelity difference} $d_{\rm fid}$ measures the expected difference in predictions of the two models over the entire feature distribution. That is, 
{
\begin{equation*}
    d_{\rm fid}(f_\theta, f_{\theta'}) = \mathbb{E}_{\mathbf{x} \in \mathcal{D}_{F}}[{\rm dist}(f_\theta(\mathbf{x}), f_{\theta'}(\mathbf{x}))],
\end{equation*}
}
where the distance function ${\rm dist}(f_\theta(\mathbf{x}), f_{\theta'}(\mathbf{x}))$ is defined based on the target model and attack goals.
\end{definition}
Label agreement~\cite{jagielski2020high} is a commonly used distance function in model fidelity attacks on classification models, where
${\rm dist}(f_\theta(\mathbf{x}), f_{\theta'}(\mathbf{x})) = \mathbbm{1}(\arg\max(f_\theta(\mathbf{x})) \neq \arg\max(f_{\theta'}(\mathbf{x})))$.
Besides label agreement, other distance metrics have been proposed to measure the fidelity difference between the target model and the reconstructed model. Chen~\etal~\cite{chen2021stealing} proposed the fidelity metric for deep reinforcement learning using the Jensen-Shannon (JS) divergence of the action probability distributions between the reconstructed model and the target model. In attacks on model architectures, attack fidelity can also be measured as the mean square error (MSE) or edit error of the reconstructed architectural parameters~\cite{batina2019csi, hu2020deepsniffer}. 

Measuring fidelity over the entire feature distribution $\mathcal{D}_{F}$ can be challenging. Therefore, the evaluations for most model fidelity extraction attacks are performed over the test datasets of the underlying learning task~\cite{jagielski2020high, rolnick2020reverse, shen2022model}. In addition, Carlinin~\etal~\cite{carlini2020cryptanalytic} measured the efficacy of the attack by synthesizing a billion queries over the entire input space.

\end{document}